\documentclass[letterpaper]{article} 
\usepackage{aaai2026}  
\usepackage{times}  
\usepackage{helvet}  
\usepackage{courier}  
\usepackage[hyphens]{url}  
\usepackage{graphicx} 
\usepackage{natbib}  
\usepackage{caption} 
\usepackage{tabularx}
\usepackage{algorithm}
\usepackage{listings}
\usepackage{algpseudocode}

\usepackage{newfloat}
\usepackage{listings}
\DeclareCaptionStyle{ruled}{labelfont=normalfont,labelsep=colon,strut=off} 
\floatstyle{ruled}
\newfloat{listing}{tb}{lst}{}
\floatname{listing}{Listing}
\title{Trustworthy AI in the Agentic Lakehouse: from Concurrency to Governance\\ }
\author {
     Jacopo Tagliabue\thanks{Corresponding author.}\textsuperscript{\rm 1},
    Federico Bianchi\textsuperscript{\rm 2},
    Ciro Greco\textsuperscript{\rm 1},
}
\affiliations {
    \textsuperscript{\rm 1}Bauplan Labs\\
    \textsuperscript{\rm 2}Together AI\\
    jacopo.tagliabue@bauplanbs.com, federico@together.ai, ciro.greco@bauplanbs.com
}
\begin{document}

\maketitle

\begin{abstract}
Even as AI capabilities improve, most enterprises do not consider agents trustworthy enough to work on production data. In this paper, we argue that the path to trustworthy agentic workflows begins with solving the infrastructure problem first: traditional lakehouses are not suited for agent access patterns, but if we design one around \textit{transactions}, governance follows. In particular, we draw an operational analogy to MVCC in databases and show why a direct transplant fails in a decoupled, multi-language setting. We then propose an agent-first design, \texttt{Bauplan}, that \textit{reimplements} data and compute isolation in the lakehouse. We conclude by sharing a reference implementation of a self-healing pipeline in \texttt{Bauplan}, which seamlessly couples agent reasoning with all the desired guarantees for \textit{correctness} and \textit{trust}.
\end{abstract}

\section{Introduction}
The lakehouse is the enterprise standard for data and AI workloads in the cloud~\cite{Zaharia2021LakehouseAN}. Notwithstanding the steady progress in coding and tool usage by Large Language Models (LLMs) \cite{shen2024llm}, production deployments of AI agents have rarely, if ever, targeted lakehouse use cases: even before considering the specificity of data \textit{vs.} software engineering, the primary obstacles in industry are trust and governance: it is unclear how a human-centered lakehouse can provide the necessary guarantees for an agent-first world. To make a  vivid example, imagine empowering a code assistant with access to \textit{your} lakehouse: what if the agent drops a table? Or if it pollutes the lake with hallucinated data?

On the research side, the data management community is raising similar concerns: it is unlikely that data systems designed for small, expensive human teams can successfully adapt to the access patterns of a \textit{swarm} of cheap AI agents \cite{tagliabue2025safeuntrustedproofcarryingai}. In particular, managing agents that run queries and build data pipelines on a lakehouse means managing concurrency on a different scale than what traditional OLAP systems are designed for \cite{liu2025supportingaioverlordsredesigning}. 

In this position paper, we argue that the industry and the research concerns above are best thought of as \textit{two sides of the same coin}. In a nutshell, correct concurrent workloads require us to solve the isolation of data and compute through a unified API, reducing the governance challenge to the well-known pattern of API-based access control. If we focus our engineering effort on making sure multiple agents can work on the lakehouse without catastrophic consequences, we will get an easy and principled path to governance as well. 

We know that this path is not impossible because -- at the very least -- \textit{it has already happened once}; in particular, we first establish a connection with the theory of multi-version concurrency control (MVCC)~\cite{10.5555/12518} in databases. Monolithic, SQL-only transactional databases (e.g., \textit{Postgres}) have evolved as a sophisticated answer to concurrency issues -- declarative abstractions, isolated processes, data snapshots allow correctness in the face of multiple users, with RBAC access control layered on top for governance \cite{10.1145/501978.501980}. The lessons from the data management field are precious, but a direct mapping of the existing techniques will \textit{not} work in a distributed, heterogeneous system such as a lakehouse.  Alas, not all hope is lost. We describe novel abstractions and isolation primitives for an agent-first lakehouse \texttt{Bauplan}. We show how to accommodate the usage patterns of swarms of agents, and easily derive the required governance: a concise, narrow API surface is much easier to protect as opposed to the plethora of tools in traditional lakehouses.

This position paper is organized as follows: Section~\ref{sec:concurrency} introduces concurrency ideas from the MVCC literature, to provide a concrete, extremely successful mental model that guides our search for similar lakehouse primitives. In Section~\ref{sec:lakehouse}, we show that naively mapping MVCC concepts won't work, before describing our proposal for lakehouse-native concepts in Section~\ref{sec:Bauplan}. We conclude with an agentic, open-source implementation of a complex data engineering task: as the landscape is evolving quickly, we do believe that a working system is a valuable reference for practitioners coming from both the agentic and the data management field; our principles, however, are indeed tool-agnostic.

\section{The MVCC mental model}
\label{sec:concurrency}

Borrowing the metaphor from \citet{10.5555/3299537}, we could think of the job of a database as giving each user the ``illusion'' of being the only user in play, while juggling safely and transparently the workloads of many such users. Correctness in the presence of concurrency is handled through \textit{transactions}~\cite{10.5555/12518}: readers only see a consistent view of the data, and writers atomically publish all the changes or, in case of error, none of them. For the current purposes, we carry along the entire paper a three-way partition of the conceptual space: data isolation, compute isolation and programming abstractions.

\begin{figure}
    \centering
    \includegraphics[width=\linewidth]{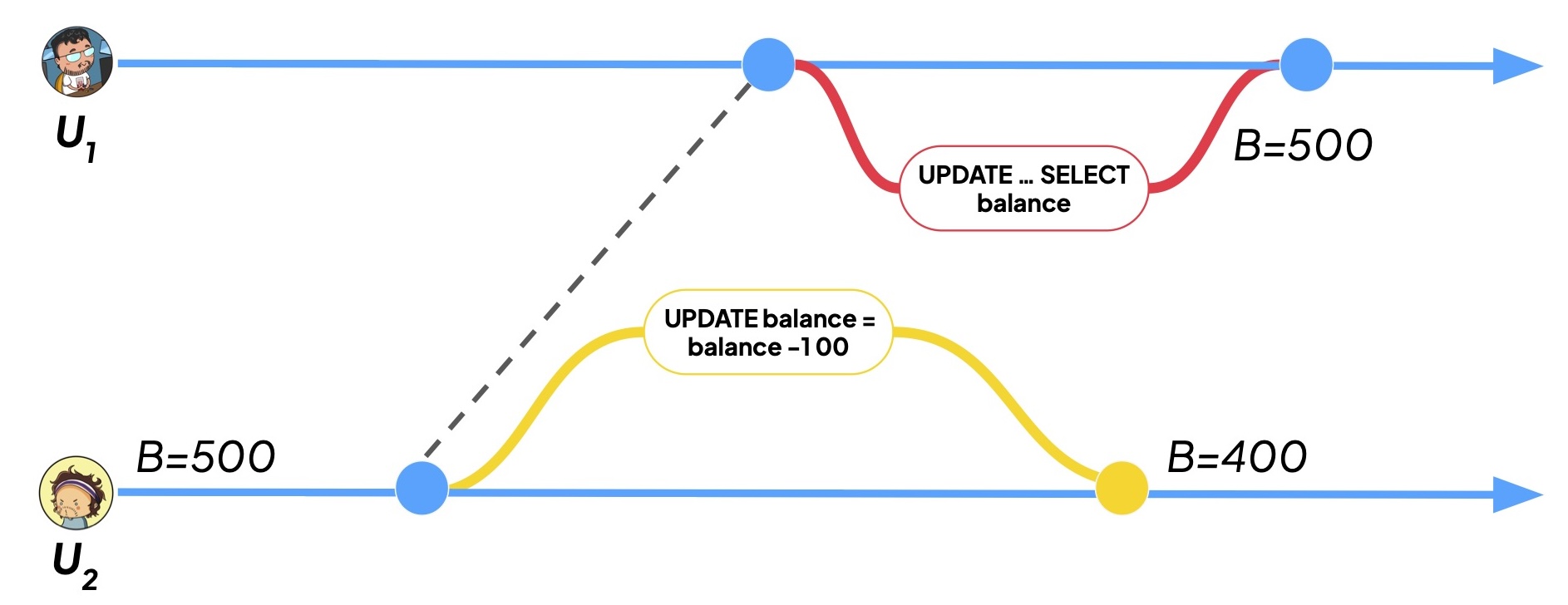}
    \caption{\textbf{The MVCC mental model}: $U_1$ starts his transaction, which at the end returns the value of \(B\) -- effectively, his code works \textit{as if} $U_2$'s transaction had never happened.}
    \label{fig:mvcc}
\end{figure}

\subsection{Data Isolation}
While the exact details are complex \cite{cerone_et_al:LIPIcs.CONCUR.2015.58}, the basic model of data isolation is straightforward. Inside of transaction boundaries, processes read data always \textit{as if} the database was frozen at the start; in order to do so, the system needs to maintain a reference to the state of the data at any point in time (a \textit{snapshot}). In Fig.~\ref{fig:mvcc}, $U_1$'s read of $U_2$'s value will return \(500\) as if $U_2$'s update never started. On the write path, transactions are used to prevent incorrect updates and race conditions: two transactions won't both commit conflicting writes to the same key based on stale reads. Importantly, none of these storage-level implementations is exposed to the final user -- a crucial design point to which we will return to below.

\subsection{Compute Isolation}

Each transaction gets processed in isolation, with no data sharing in between -- from the point of view of the single transaction, it is \textit{as if} that is the only process active at that time. In a SQL database there is no equivalent of package management, resulting in a small toolchain as opposed to a more heterogeneous compute environment where isolation is not as obvious.

\subsection{Programming Abstractions}

Traditional databases leverage a declarative language such as SQL which simplifies the relationship between the system and its users. Users express \textit{what} data they wish to read or write, the query engine and the storage engine figure out \textit{how} that should happen. This is a crucial design choice for at least two independent reasons:

\begin{itemize}
    \item \textit{correctness}: the complexity of coupling data and compute isolation within transactions is offloaded entirely to the platform, avoiding the common pitfalls of \textit{ad hoc} transactions that plague even popular open source projects \cite{10.1145/3638553};
    \item \textit{performance}: while each user is given the illusion of being alone, the platform may re-use transparently work done for one user for another one (e.g., caching).
\end{itemize}

A final consequence deserves a special mention,     \textit{trust}. Since the only way to interact with the system is at the logical layer, protecting the physical representation of the data is now as easy as implementing RBAC over users -- unauthorized code does not even go beyond the planning module, ensuring no contamination at the physical layer.

As we shall see in the ensuing section, the properties that make MVCC workable in a monolithic database -- uniform runtime, co-located control over data and execution -- do not hold in lakehouses with heterogeneous runtimes and decoupled storage, so a direct transplant is insufficient.

\section{From MVCC to the lakehouse}
\label{sec:lakehouse}

A transactional database is a vertically integrated software, which controls storage, compute and APIs. A lakehouse is a distributed, multi-language system built on the principle of storage-compute decoupling. The most typical workload is a data pipeline~\cite{Renen2024}, a DAG of transformations going from raw data in source tables to intermediate and final assets for downstream consumption (e.g., a dashboard, a RAG system). As it turns out, adapting MVCC principles to this architecture is not straightforward.

\subsection{Data Isolation}

The storage layer of a lakehouse is typically built on Apache Iceberg tables, which guarantee transaction-like behavior on a \textit{per-table} basis. However, single table guarantees are not enough when most writes are pipelines: if an agent writes wrong code at node \(3\) of a pipeline, nodes \(1\) and \(2\) are individually sound transactions, but the lake is in an inconsistent state: downstream readers have no protection, they may \texttt{JOIN} a new version of \(2\) with an old version of \(3\). We will come back to this distinction when discussing Fig.~\ref{fig:transaction} below.

\subsection{Compute Isolation}
\label{sec:compIsolation}

A data pipeline with three nodes may require SQL for the first one, Python \(3.10\) with \texttt{pandas} for the second, and Python \(3.11\) with \texttt{polars} for the third one. In other words, a lakehouse must support truly heterogeneous compute: on the one hand, building and running general purpose Python code requires a novel platform-level toolchain; on the other, Python code introduces new vulnerabilities, such as installing malicious packages or accessing the internet. Traditional lakehouses have historically dealt with heterogeneous use cases by \textit{adding} runtime options and execution workflows
\cite{cdms2025eudoxia}: today, it is common to move between different GUIs, APIs and interfaces to master the full lifecycle of a single pipeline. Unlike in the MVCC case, compute isolation is anything but obvious.

\subsection{Programming Abstractions}
\label{sec:airflow}

Heterogeneous compute makes it harder to tie execution back to a consistency model (e.g., how do I reconcile a warehouse \texttt{INSERT} with a Spark job?). We can highlight the issue with scattered abstractions discussing mainstream reference implementations. Let's consider the \textit{AWS} example for a data pipeline using Airflow \cite{awsairflow}\footnote{Note that an equivalent example in the data engineering world can be found in the Databricks Best Practices with Scala \cite{dx}.}:

\begin{lstlisting}[
  language=Python,
  showstringspaces=false,
  columns=fullflexible,
  caption={A pre-processing node in an Airflow DAG},
  basicstyle=\ttfamily\scriptsize,
  numbers=none
]
def preprocess(s3_in_url, s3_out_bucket, s3_out_prefix):
    # Do pre-processing and save the result in
    # "s3_out_bucket / s3_out_prefix"
    return "SUCCESS"

preprocess_task = PythonOperator(
    task_id="preprocessing",
    dag=dag,
    python_callable=preprocess.preprocess
    )
\end{lstlisting}

In this paradigm, we lost two properties we care about:

\begin{itemize}
    \item \textit{correctness}: there are no transactional guarantees anymore as the user is in charge both of the storage layer and the DAG process. It is therefore impossible to handle simple failures in a principled way: e.g., what is the state of the system if the process fails just before \texttt{return}?
    \item \textit{performance}: the function writes its output  as a side effect, preventing further optimizations on the outer loop (e.g., the DAG process cannot easily cache results).
\end{itemize}

The independence between compute (the DAG process) and data (the physical files in S3) has immediate implications for trust and governance: to be able to iterate over this DAG, an agent should be given access to i) the physical layer for the files, ii) Python-specific infrastructure to invoke the runtime, iii) and to a SQL engine if it needs to perform exploratory queries when debugging. As the surface expands, governance becomes \textit{exponentially} more difficult.

\section{The Agentic Lakehouse}
\label{sec:Bauplan}

Naive application of the MVCC model to a distributed system will fail to achieve the goal of correctness-under-concurrency that we are pursuing for the agentic lakehouse, because physical decoupling and heterogeneous runtimes change the isolation contract. Not all is lost, though: we now describe \texttt{Bauplan} as a concrete implementation of those concepts, designed from first principles for the lakehouse. 

\subsection{Data Isolation}

If every \textit{write} in the lakehouse is immutably recorded, a series of snapshots track the history of tables through time. From this simple primitive, it is trivial to quickly model concurrent changes to \textit{tables} in the same way \textit{Git} models concurrent changes to \textit{code}  \cite{10.1145/2661136.2661137} -- a snapshot is identified by a \textit{commit} with a parent, the \texttt{HEAD} of a sequence of commits is a \textit{branch} and resolving possible conflicts between two branches happens within a \textit{merge}. \texttt{Bauplan} adopts an efficient, copy-on-write branching mechanism, so that branching can be done efficiently even on hundreds of tables and billions of rows. The crucial insight is once again related to the nature of data workloads: as pipelines are multi-table, they would span \textit{multiple commits}. These abstractions can now model transaction boundaries across arbitrarily long DAGs by decoupling single table writes (commits) and multi-table atomic changes (merges) (Section~\ref{sec:transactions}).

\subsection{Compute Isolation}

As any lakehouse workload can be built out of functions with different latency constraints \cite{cdms2025eudoxia}, 
\texttt{Bauplan} adopts a Function-as-a-Service (FaaS) model as the underlying unified compute \cite{10.1145/3702634.3702955}. The FaaS model perfectly aligns with our compute isolation needs -- each function runs its own (containerized) Python / SQL engine, independent of \textit{any} other function and completely \textit{isolated from the public internet} -- and with a division of labor where agents provide code, and hardware, operating system and runtime are under the platform's purview \cite{196322}.

\subsection{Programming Abstractions}
\label{sec:transactions}

A FaaS model maps immediately to DAG abstractions that are \textit{functional}. Agents can code a DAG by chaining functions together: 

\begin{lstlisting}[
  language=Python,
  showstringspaces=false,
  columns=fullflexible,
  caption={Declarative I/O and infra-as-code},
  basicstyle=\ttfamily\scriptsize,
  numbers=none
]
@bauplan.model(materialization='REPLACE', name='A')
@bauplan.python('3.10', pip={'pandas': '2.0'})
def parent(
    trips = bauplan.Model("taxi_trips"),
    zones = bauplan.Model("taxi_zones")
):
    # business logic only; 
    # I/O is platform-mediated
    return trips.join(zones).do_something()

@bauplan.model(materialization='REPLACE', name='B')
@bauplan.python('3.11', pip={'polars': '0.88'})
def child(df = bauplan.Model("parent")):
    return df.do_something()
\end{lstlisting}

The proposed abstractions take the typical infrastructure-as-code approach of industry FaaS and simplify it further by expressing declarative environments through a Python decorator LLMs can easily recognize. When compared with the code from Section~\ref{sec:airflow}, the second major difference is declarative I/O -- functions accept tables (not files) as input, and output tables directly.

\begin{figure}
\centering
\includegraphics[width=\linewidth]{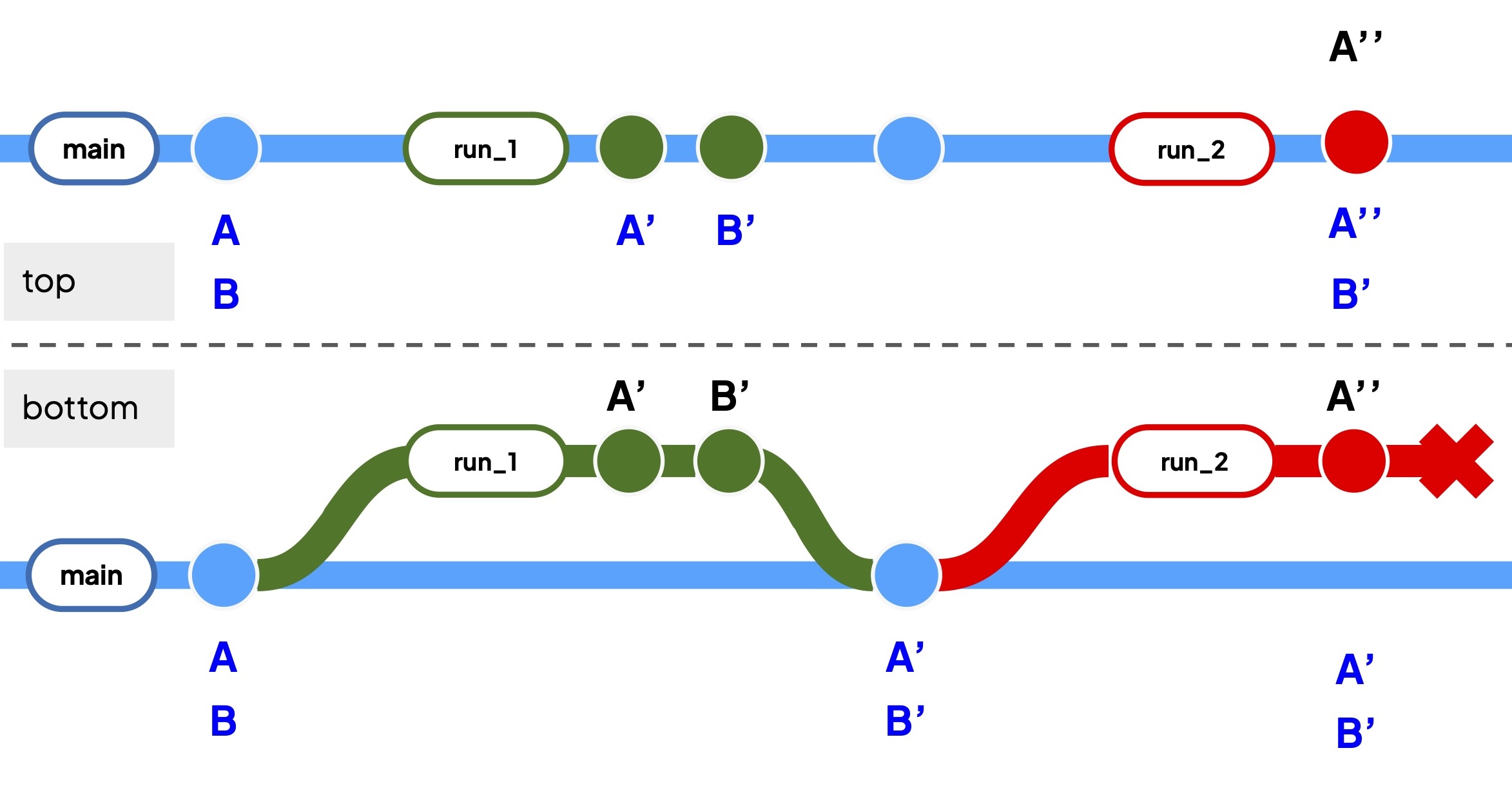}
\caption{\textbf{Transactional pipelines.} \textit{Top}: without coupling temporary branches with pipeline runs, \texttt{run\_2} will leave in \textit{main} a new version of  \(A\) but an old version of \(B\). \textit{Bottom}: \texttt{Bauplan} \textit{run} API will guarantee atomic write of \(A'\) and \(B'\) on success -- \texttt{run\_1} --, and isolation in case of failure -- \texttt{run\_2}.}
\label{fig:transaction}
\end{figure}

As compute and storage are decoupled, we need one more API to \textit{bind} together declarative pipelines, FaaS execution and data branches. This is easily accomplished with a unified API to launch all workloads, i.e. \texttt{bauplan.run(my\_pipeline)}. During a \textit{run} on the \textit{main} data branch, \texttt{Bauplan} will: \(1)\) automatically open a \textit{temporary} branch; \(2)\) fetch from S3 the rows corresponding to the DAG source tables; \(3)\) run user code on the FaaS runtime and perform the required \textit{writes} to S3; \(4)\) \textit{on success}, merge the temporary branch to \textit{main}; \textit{on failure}, leave the temporary branch open, and \textit{main} untouched. Extending declarative abstractions to Python, and achieving data-compute logical unification (even with a physical decoupling) get us back the desired properties:

\begin{itemize}
    \item \textit{correctness}: the platform supports transactional pipelines across languages and an arbitrary amount of tables. Figure~\ref{fig:transaction} illustrates the importance of conceiving transactions as an abstraction that embraces data \textit{and} compute. Without enforcing temporary branches (Figure~\ref{fig:transaction}, top), a two-node pipeline failing after one node leaves \textit{main} in an inconsistent state \textit{even if} the single-table transaction was successful;
    \item \textit{performance}: declarative abstractions leave the platform free to optimize execution \textit{across agents} leveraging its privileged, centralized position.
\end{itemize}

\subsection{From concurrency control to governance}

As highlighted in Section~\ref{sec:lakehouse}, heterogeneous compute opens the door for agents to run untrusted logic. We are now in a position to tie back together our abstractions to trust and governance:

\begin{itemize}
    \item \textit{data governance}: on the \textit{read} path, declarative I/O provides a unique SQL-like layer to check for authorization; on the \textit{write} path, the \textit{merge} primitive allows a fail-safe mechanism that safeguards production data;
    \item \textit{compute governance}: as functions are fully isolated and do not require internet access, packages remain the only major vector of attack. Once again, the declarative APIs make it trivial to enforce rules at the platform level by checking a decorator against a whitelist: no agent is able to install dangerous packages;
    \item \textit{API governance}: as all workloads are invoked through a simple set of unified APIs, securing access boils down to tried-and-tested patterns, such as RBAC with fine-grained permissions over data operations and executable units.
\end{itemize}

Solving concurrency and correctness is the key to unlocking an easy path to governance and trust: the agentic lakehouse is indeed a re-implementation of the successful MVCC patterns in the distributed lakehouse world.

\subsection{A worked-out example: self-healing pipelines}
\label{sec:example}

\begin{figure}
    \centering
    \includegraphics[width=\linewidth]{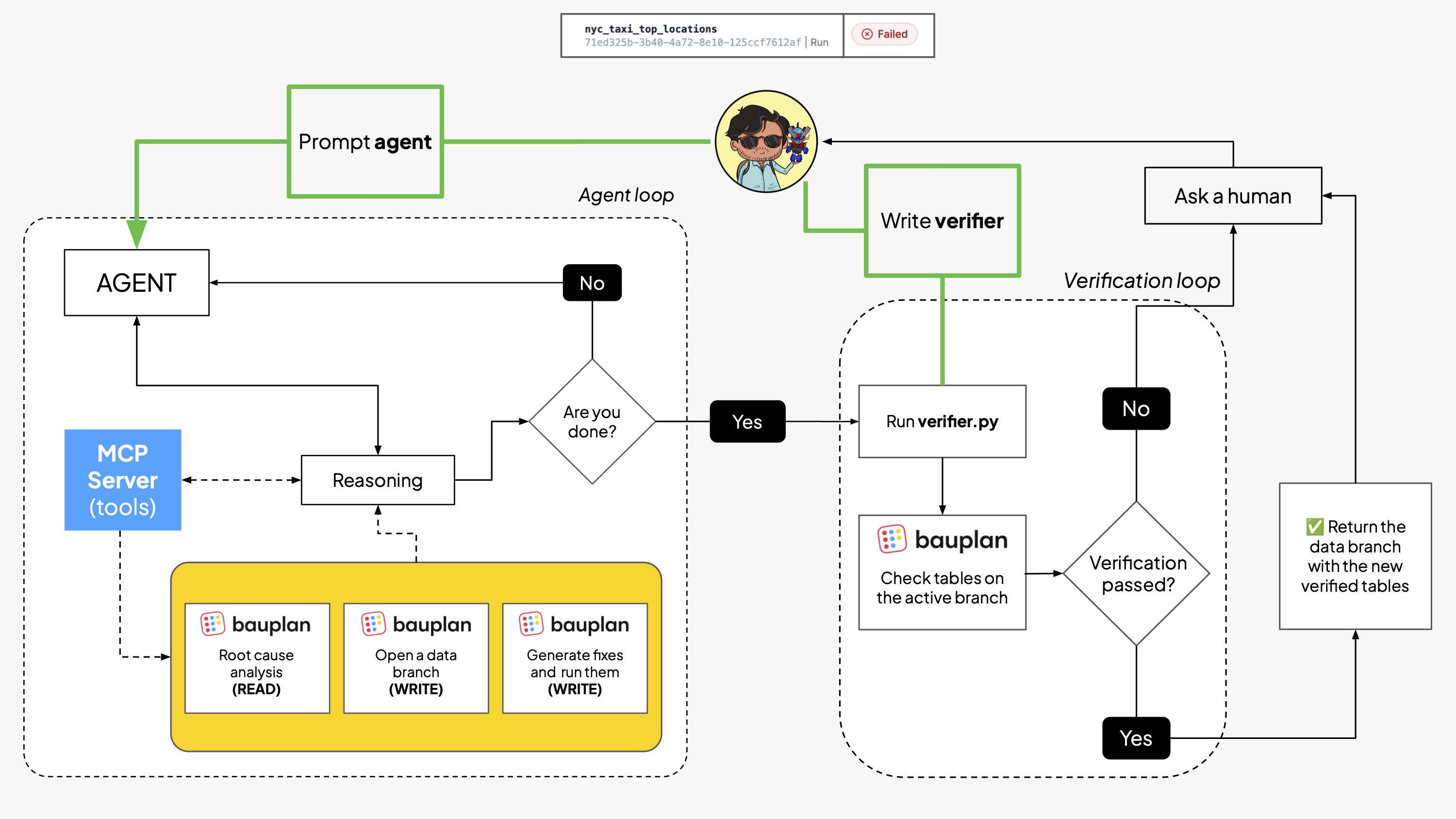}
    \caption{\textbf{Self-healing pipelines}: a ReACT loop is triggered on the agentic lakehouse -- at the end, a verifier acts as a first sanity check on the agent work, leaving to a human the final confirmation thanks to the \textit{branch-then-merge} flow.}
    \label{fig:agent}
\end{figure}

As an example of a complex and very important use case that can be solved safely in the agentic lakehouse, we propose the implementation of a self-repairing pipeline.\footnote{https://github.com/BauplanLabs/the-agentic-lakehouse} Fig.~\ref{fig:agent} illustrates the high-level flow: a run failed, triggering an agentic resolution. A human engineer will prompt the agent for a resolution and produce a \textit{verifier}, i.e. computational ``acceptance criteria'' registered within the platform itself. A ReAct loop \cite{yao2023reactsynergizingreasoningacting} of reasoning over \texttt{Bauplan} APIs comprises the bulk of the agentic work, leveraging the abstractions above: code execution is sandboxed \textit{and} every \textit{write} happens on a data branch, leaving production safe and untouched -- similarly to what happens with code reviews, the agent's output is a (data) \textit{branch}. The Git-like primitives provide a natural hook for human-in-the-loop verification: first, the platform runs the verifier to independently establish that minimal criteria are met; second, the review-then-merge flow happens with the supervision of an engineer --- on success, copy-on-write merge will ensure efficient publication of the agent's work into \textit{main}.

\section{Conclusion and future work}
\label{sec:conclusion}
We advocate correctness by construction in an agent-first lakehouse: declarative I/O, temporary branches with atomic merge, and network-isolated functions bound to a single \texttt{run} API. As a concrete implementation of these principles, we described the agentic lakehouse \texttt{Bauplan} and argued that pursuing safe concurrency in a lakehouse creates a clear path for principled governance. The bottleneck to scale trustworthy data engineering output is \textit{not} intelligence anymore, but infrastructure.



\bibliography{aaai2026}

\end{document}